\documentclass[10pt,journal,cspaper,compsoc]{IEEEtran}
\usepackage{verbatim} 
\usepackage{algorithmic}
\usepackage{graphicx}
\usepackage{caption}
\usepackage{subcaption}
\usepackage{bbm, dsfont}
\usepackage{caption}
\usepackage{multirow}
\usepackage{amssymb,amsmath}
\usepackage{footnote} 
\usepackage{breqn}
\usepackage{soul}
\usepackage{flushend}

\usepackage{tikz}
\usetikzlibrary{fit,positioning}
\usepackage{xcolor}
\usetikzlibrary{patterns}

\ifCLASSOPTIONcompsoc
\else
\fi
\ifCLASSINFOpdf
\else
\fi
\begin{document}
%
\title{Parsimonious Topic Models \\with Salient Word Discovery}
%
%
%
%

\author{Hossein~Soleimani,
        and~David~J.~Miller,~\IEEEmembership{Senior Member,IEEE}
\IEEEcompsocitemizethanks{\IEEEcompsocthanksitem H. Soleimani (email: hsoleimani@psu.edu) {and} D. J. Miller (email: djmiller@engr.psu.edu) are with the Department of Electrical Engineering, Pennsylvania State University, University Park, PA, 16802.}
\thanks{}}

%
%

\markboth{}%
{Shell \MakeLowercase{\textit{et al.}}: Bare Demo of IEEEtran.cls for Computer Society Journals}
%


\IEEEcompsoctitleabstractindextext{%
\begin{abstract}
We propose a parsimonious topic model for text corpora. In related models such as Latent Dirichlet Allocation (LDA), all words are modeled topic-specifically, even though many words occur with similar frequencies across different topics. Our modeling determines salient words for each topic, which have topic-specific probabilities, with the rest explained by a universal shared model. Further, in LDA all topics are in principle present in every document. By contrast our model gives sparse topic representation, determining the (small) subset of relevant topics for each document. We derive a Bayesian Information Criterion (BIC), balancing model complexity and goodness of fit. Here, interestingly, we identify an effective sample size and corresponding penalty specific to each parameter type in our model. We minimize BIC to jointly determine our entire model -- the topic-specific words, document-specific topics,  all model parameter values, {\it and} the total number of topics -- in a wholly unsupervised fashion. Results on three text corpora and an image dataset show that our model achieves higher test set likelihood and better agreement with ground-truth class labels, compared to LDA and to a model designed to incorporate sparsity.
\end{abstract}

}

\IEEEaftertitletext{\vspace{-1.8\baselineskip}}
\maketitle

\IEEEdisplaynotcompsoctitleabstractindextext

%
\IEEEpeerreviewmaketitle

\section{Introduction}
%
%

%
%
%
%

\IEEEPARstart{P}{arsimonious} models 
are economical in the number of free parameters they require to model data.
Such models have been proposed {\it e.g.} for 
model-based clustering, reducing the number of free (covariance matrix) parameters 
needed to
specify a cluster's shape (which, without parsimony, grows quadratically with the number of feature dimensions) \cite{Fraley1998}, \cite{Ghahramani1999}.  For clustering of text documents, great reduction in the number
of free parameters has also been achieved by {\it tying} parameter values across many clusters, {\it e.g.} \cite{Graham2006}.
Parsimonious models
are less prone to overfitting
than parameter-rich ones in various modeling problems \cite{Jefferys1992},\cite{MacKay2003}. Document modeling is a good ``target'' here, as the ``bag-of-words'' model \cite{Manning1999} introduces one parameter per word, per topic, for each word in a given lexicon. This can amount to tens of thousands of word probability parameters for \textit{every} topic in the model. It is expected that economization on this model representation should be possible. 

In topics models such as Latent Dirichlet Allocation (LDA) \cite{Blei2003}, each topic is a probability mass function defined over the given vocabulary. LDA is an extension of mixtures of unigrams, with the latter assuming each document is generated by a single topic randomly selected based on corpus-level topic proportions \cite{Nigam2000}. By contrast, 
in LDA, each word in a given document is independently modeled by a (document-specific) mixture over the topics.

In LDA, each word has a freely chosen probability parameter under every topic. This entails a huge set of parameters to estimate and hence makes the model prone to over-fitting. Moreover, intuitively, many words are not context-specific, 
i.e. they may be used with roughly the same frequency under different topics. Thus, these (many) words may be
reasonably modeled using a universal {\it shared model} across all topics.  Our approach will exploit such parsimony. 

Similarly, in LDA every topic is in principle present in every document, with a non-zero proportion. This seems implausible since each document is expected to have a main theme covered by a modest subset of related topics. Allowing all topics to have nonzero proportions in every document again complicates the model's representation of the data.  By contrast, our proposed method identifies a sparse set of topics present in each document.

Our model is in fact sparse in both abovementioned senses. First, under a given topic, each word may be a topic-specific feature with its own probability parameter or a shared feature, using a parameter shared with other topics. Second, for each document a sparse set of ``occurring'' topics -- those with non-zero proportions --  is identified.
These two sparsities can be understood from Fig. \ref{fig1}, with $\beta_{jn}$ the probability of topic $j$ generating word $n$
and $\alpha_{jd}$ the proportion of topic $j$ in document $d$. For example, here only topics 1 and 3 are present in document 1, and
the first word's probability of generation is shared by all the topics. 

Our model learning consists of two main parts, alternately, iteratively applied: structure learning and parameter learning. Assuming the number of topics is known and given current parameter estimates, {\it structure} learning determines the topic-specific words for each topic and the topics with non-zero proportions in each document. Likewise, given a fixed structure, in parameter learning the document-specific topic proportions and topic-specific word probabilities are re-estimated. In the following sections we introduce an objective function and a computationally efficient procedure, jointly performing structure and parameter learning to minimize this objective. 
\begin{figure}[t]
		\begin{tikzpicture}[scale=.35,shift={(-9.2,-5.5)}]
		\node[] at (15,3) [label=left:\mbox{\footnotesize $\beta=\begin{bmatrix}
  \mathbf{0.1}&0.2&\mathbf{0.2}& 0.25 &0.1&\mathbf{0.15}\\
  \mathbf{0.1}&0.25&\mathbf{0.2}&0.3&0.05&0.1 \\
  \mathbf{0.1}&0.05&0.3&0.1&\mathbf{0.3}&\mathbf{0.15} \\
 \end{bmatrix}$}] {};
 	\node[] at (22,3) [label=left:\mbox{\footnotesize $\alpha=\begin{bmatrix}
  0.2 & 0 \\
  0 & 0.4 \\
  0.8 & 0.6\\
 \end{bmatrix}$}] {};
 \end{tikzpicture}                
\caption{A simple example to illustrate sparsity on word probabilities ($\beta$) (left) and topic proportions ($\alpha$) (right) for a corpus with 2 documents, 6 words, and 3 topics. (Shared words are in bold.)}
\vspace{-0.2in}
        \label{fig1}
\end{figure}

Our objective function is derived from the model posterior probability \cite{Kass1995},\cite{Bishop2006}. One such criterion is the Bayesian Information Criterion (BIC) \cite{Schwarz1978}. BIC, which is a widely accepted criterion for model comparison, is the negative logarithm of the Bayes marginal likelihood and is composed of two main terms: the data log-likelihood and the model complexity cost.  BIC thus achieves a balance between goodness of fit and model complexity. 

However, BIC has some deficiencies which limit its applicability. The parameter penalty in BIC ($\frac{1}{2}\log(sample~size)$) is the same for all parameters in the model \cite{Schwarz1978}, while we expect different parameter types in general contribute unequally both to model complexity and to goodness of fit. Moreover, the Laplace's approximation used in deriving BIC is only valid when 
the ratio of sample size to feature space dimensionality is large
\cite{Kass1995}. However, in document modeling this ratio is very small because the
feature dimensionality is the (huge) dictionary size.

In this paper, we derive an approximation of the model posterior which improves on the na\"{\i}ve form of BIC in two aspects: 1) Our proposed form of BIC has differentiated cost terms, based on different \textit{effective} sample sizes for the different parameter types in our model. 2) Making use of a shared feature representation essentially increases the sample size to feature dimension ratio, 
thus giving a better approximation of the model posterior.

Our framework also gives, in a wholly unsupervised fashion, a direct estimate of the number of topics present in the corpus. The number of topics (i.e. model order) is a hyper-parameter in topic models, usually determined based on validation set performance for a secondary task such as classification.  Here, solely using the (unlabeled) document corpus, we select the model order with highest approximate model posterior (minimum BIC). Thus,
the model in its entirety -- its structure, parameter values, {\it and} the number of topics -- are all jointly
chosen to minimize BIC.

\subsection{Related Work}
``Common'' and ``specific'' words, sparsity in topic proportions and word probabilities, as well as estimation of the number of topics have all been the subject of previous studies. \cite{Wallach2009a} introduced asymmetric Dirichlet priors over topic distributions and word probabilities to control skewness in word and topic frequency distributions. Asymmetric priors were shown to prevent common words from dominating all topics and also help achieve sparser topic presence in documents. However, similar to LDA, this approach is not parsimonious. All topics have nonzero proportion in every document and all words are modeled in a topic-specific fashion.
\cite{Wang2009} introduced a spike and slab model to control sparsity in word probabilities. Unlike our approach, \cite{Wang2009} does not use a shared distribution. Moreover, it does not provide the subset of relevant topics for each document. A similar approach, based on the Indian Buffet Process, was used in \cite{Williamson2010} to address sparsity only in topic proportions in a non-parametric topic model.

Global background models have been used in information retrieval \cite{Street1998}, \cite{Zhang2002}, \cite{Zhai2001}. In these models, the probability of each word under every topic is a mixture of the background model and topic-specific word probabilities. A similar idea has been used in \cite{Hiemstra2004}, with words well-modeled by the background model having small (near zero) topic-specific probabilities. The mixing proportions in these models are hyperparameters that should be estimated by cross-validation. \cite{Chemudugunta2007} proposed a combination of background, general, and document-specific topics to improve information retrieval. The authors argued that LDA ``over-generalizes'' and is thus not effective for matching queries that contain both high-level semantics and keywords. \cite{Chemudugunta2007} introduced a huge set of new free parameters by adding a document-specific topic for every document. 
In these models, similar to LDA, each word, under every topic, has a free parameter. By contrast, in our model, the shared model is heavily used, with each topic possessing relatively few topic-specific words. Also unlike these approaches, our model is sparse in topic proportions.

\cite{Zhu2011} presented Sparse Topical Coding (STC), a non-probabilistic topic model which gives parsimony in topic-proportions but models all words topic-specifically. Moreover, this method has three hyper-parameters that must be determined by cross-validation. 

Non-parametric topic models have been proposed that relax the requirement of specifying the number of topics \cite{Teh2006b}. However similar to LDA, these methods do not exhibit parsimony in their modeling.

Our approach can also be viewed from the standpoint of unsupervised feature selection. For each topic we select salient features in an unsupervised fashion, modeling the rest using a universal shared model. 
Unsupervised feature selection and shared feature representations have been considered in some prior works. 
\cite{Law2004} used a minimum message length criterion to find salient features in a mixture model. Features were tied across all components; i.e. each feature is either salient or shared in all components. A related Bayesian framework was presented in \cite{Constantinopoulos2006}. 

The concept of shared feature space for mixtures was further improved in \cite{Graham2006} by proposing a component-specific feature space; i.e. a feature can be salient in some components but represented by the shared model in others. \cite{Graham2006} used the Minimum Description Length (MDL) \cite{Grunwald2007} and standard mixture of unigrams for modeling documents. \cite{Boutemedjet2009} performed unsupervised feature selection by minimizing the message length of the data, considering mixtures of generalized Dirichlet distributions. This model was then optimized in a Bayesian framework via variational inference in \cite{Fan2013}. 

\textit{Contributions of this paper:} Compared to previous works, our main contributions are:
\begin{enumerate}
\item We extend the concept of shared feature space from standard mixtures to more general topic models, allowing presence of multiple topics in documents. In doing so, we achieve sparsity in topic proportions and in topic-specific words.  Prior works at best achieve sparsity in one of these two senses.
\item Unlike most works, our model allows the subset of salient words to be topic-specific. This follows the premise that some words may have common frequency of occurrence under some subset of the topics. For example, the word ``component'' has different meanings under ``statistics'' and ``machine learning'' than under other topics, and could have higher frequencies of occurrence for these specialized topics.
\item We derive a novel BIC objective function, used for learning our model. Unlike the na\"{\i}ve form, satisfyingly, our derived objective has distinct penalty terms for the different parameter types in our model, interpretable vis a vis the {\it effective} sample size for each of the parameter types. 
\end{enumerate}

The rest of the paper is organized as follows: Section \ref{notation} introduces the notation we use throughout. In section \ref{ldarev}, we briefly review LDA and introduce a criterion to measure sparsity of topics for LDA. Next, in section \ref{ourmodel}, we present our parsimonious model. Section \ref{bicder} derives our BIC objective function. Then, in section \ref{mainalg}, we develop the joint structure and parameter learning algorithm for our model, which locally minimizes our BIC objective. Experimental results on three text corpora and an image dataset are reported in section \ref{results}. Concluding remarks are in section \ref{conc}.

\vspace{-.05in}
\section{Notation and Terminology}
\label{notation}

A corpus $\mathcal{D}$ is a collection of $D$ documents and a dictionary is a set of $N$ unique words. We index unique documents in the corpus and unique words in the dictionary by $d\in \left\{1,...,D\right\}$ and $n\in \left\{1,...,N\right\}$, respectively. Also, $j\in \left\{1,...,M\right\}$ is a model's topic index, $M$ the total number of topics.

We define the following terms specific to our modeling of each document $d$:
\begin{itemize}
	\item $L_d$ is the number of words in document $d$.
	\item $w_{id}\in \left\{1,...,N\right\},~i=1,...,L_d$ is the $i$-$th$ word in document $d$.
	\item $v_{jd} \in \{0,1\}$ indicates whether ($v_{jd}=1$) or not ($v_{jd}=0$) topic $j$ is present in document $d$.
	\item $M_d\equiv \sum\limits_{j=1}^M v_{jd} \in \left\{1,...,M\right\}$ is the number of topics present in document $d$.
	\item $\alpha_{jd}$ is the proportion for topic $j$ in document $d$.
\end{itemize}

The following quantities are specific to each topic:
\begin{itemize}
	\item $\beta_{jn}$ is the topic-specific probability of word $n$ under topic $j$.
	\item $u_{jn} \in \{0,1\}$ indicates whether ($u_{jn}=1$) or not ($u_{jn}=0$) word $n$ is topic-specific under topic $j$.
	\item $N_j \equiv \sum\limits_{n=1}^N u_{jn}$ is the number of topic-specific words in topic $j$.
	\item $\bar{L}_j\equiv \sum_{d=1}^{D}{L_d{v_{jd}}}$ is the sum of the lengths of the documents for which topic $j$ is present.
\end{itemize}

Finally, $\beta_{0n}$ is the probability of word $n$ under the shared model.

\vspace{-.1in}
\section{Latent Dirichlet Allocation}
\label{ldarev}
LDA is a generative model originally proposed for extracting topics and organizing text documents \cite{Blei2003}. Its generative process for a document $d$ is as follows: 

\begin{enumerate}
\item Choose topic proportions $\alpha_d$ from a Dirichlet distribution with parameter $\eta$, i.e. $\alpha_d \sim Dir(\eta)$.
\item For each of the $L_d$ words $w_{id}$:
\begin{enumerate}
\item Choose a topic $z_{id} \sim Multinomial(\alpha_d)$.
\item Choose word $w_{id}$ according to the multinomial distribution for topic $z_{id}$, i.e. $\left\{\beta_{z_{id}n},~n=1,...,N \right\}$.
\end{enumerate}
\end{enumerate}

\subsection{Parameter Estimation in LDA}
The main inference in LDA is computing the posterior probability of the hidden variables $\{z_{id}\}$, given a document. 
Exact inference is intractable.
Thus, approximate algorithms such as variational inference \cite{Blei2003} and Markov Chain Monte Carlo \cite{Griffiths2004} have been used. Here we briefly review mean-field variational inference.

Approximate inference in this method is achieved by obtaining a lower bound on the log-likelihood. A new family of variational distributions is defined by changing some of the statistical dependencies in the original model:
\vspace{-.15in}
\begin{align}
q(\alpha_d,z_d|\gamma^{(d)},\phi^{(d)})=q(\alpha_d|\gamma^{(d)})\prod_{i=1}^{L_d}{q(z_{id}|\phi^{(d)}_{i})},~\forall d.
\label{vardist}
\end{align}
\vspace{-.10in}

Here, $z_d$ is the set of hidden variables $\left\{z_{1d},...,z_{{L_d}d}\right\}$ and $q(\alpha_d|\gamma^{(d)})$ is a Dirichlet distribution with parameter $\gamma^{(d)}$. Also, $q(z_{id}|\phi^{(d)}_{i})$ is a multinomial distribution on the $M$ topics, with variational parameters $\phi^{(d)}_{i1},...,\phi^{(d)}_{iM}$. The values for $\gamma$ and $\phi$ for each document are determined by minimizing the Kullback-Leibler (KL) divergence between $q$ and the posterior distribution of hidden variables, which gives a lower bound on the single-document log-likelihood. Update equations for the variational parameters are:
\vspace{-.1in}
\begin{equation}
\phi_{ij}^{(d)} \propto \beta_{jw_{id}}\exp\big(\Psi(\gamma_j^{(d)})-\Psi(\sum_{j'=1}^{M}{\gamma_{j'}^{(d)}})\big) ~~~  
\end{equation}\vspace{-.25in}
\begin{equation}
\gamma_j^{(d)}=\eta_j + \sum_{i=1}^{L_d}{\phi_{ij}^{(d)}}~~~~~~~~~~~~~~~~~~~~~~~~~~~~
\label{varparams}
\end{equation}
where $\Psi(\cdot)$ is the first derivative of the log-gamma function. Also the $\phi$ parameters are constrained so that $\sum_{j=1}^{M}{\phi_{ij}^{(d)}}=1~~i=1,...,L_d,~ d=1,...,D$.

In the next step, after optimizing the variational parameters, the lower bound is optimized with respect to parameters of the model (i.e. the word probabilities under each topic and $\eta$). For the word probabilities, this minimization is achieved via the closed form updates:
\begin{align}
\beta_{jn} = \frac{\sum_{d=1}^{D}{\sum_{{i=1:w_{id}=n}}^{L_d}{\phi^{(d)}_{ij}}}}{\sum_{d=1}^{D}{\sum_{i=1}^{L_d}{\phi^{(d)}_{ij}}}},~\forall n, j.
\label{betaupdate}
\end{align}\vspace{-.1in}

There is no closed-form update for $\eta$. It is updated using Newton-Raphson \cite{Blei2003}. This process is iteratively repeated until a termination criterion is met.\vspace{-.15in}

\subsection{Sparsity in LDA}
Sparsity of topic proportions is controlled by $\eta$, a corpus-level parameter optimized along with all other parameters of the model. Values of $\eta$ smaller than $1$ lead to sparser topic proportions in a given document. 

In order to compare sparsity in LDA with our model, we estimate the actual number of LDA topics present in each document. The variational parameter $\phi_{ij}^{(d)}$ is essentially the probability that word $i$ in document $d$ is generated by topic $j$. To estimate the number of topics present, we hard-assign each word in a document to the topic that has maximum $\phi_{ij}^{(d)},~j=1,...,M$. In so doing, we determine the set of topics used to model at least one word in a given document. This is how we measure topic sparsity for LDA.

In the next section, we develop our new topic model, which fundamentally differs from LDA in:
1) possessing two types of sparsities; 2) treating parameters as deterministic unknowns, rather than random variables,
i.e. we take a maximum likelihood rather than Bayesian learning approach \cite{Duda2012}; thus, our method does not require
variational inference;
3) requiring {\it no} hyperparameters.  In LDA, $M$ is the sole hyperparameter.  However,
in our approach,
$M$ is automatically estimated, along with the rest of our model.\vspace{-.15in}
\section{Parsimonious Topic Model}
\label{ourmodel}
In this section we develop our parsimonious model and its associated parameter estimation. For clarity's sake, in this section, we will focus only on parameter estimation given the model \textit{structure} assumed known, i.e. the subset of topics present in each document (specified by $\mathbf{v}=\{v_{jd}\}$), the topic-specific words under each topic (specified by $\mathbf{u}=\{u_{jn}\}$), and the number of topics, $M$. Structure learning and model order selection will be developed in section \ref{mainalg}. 

We treat model parameters as deterministic unknowns, to be estimated by maximum likelihood. While the Bayesian setting is a natural alternative, our approach (developed in section \ref{bicder}) approximates the Bayesian model posterior. Moreover, unlike a fully Bayesian approach, our approach 
gives a computationally tractable way 
for learning parsimonious models, and thus for avoiding overfitting.

\noindent
{\it Stochastic Data Generation:}

\noindent
The document corpus is generated under our model as follows.

\begin{enumerate}
\item For each document $d=1,\ldots,D$
\item For each word $i=1,\ldots,L_d$
\begin{enumerate}
\item Randomly select a topic based on the probability mass function (pmf) $\{\alpha_j v_{jd}, j=1,\ldots,M\}$.
\item Given the selected topic $l$, randomly generate the $i$-th word based on the topic's
pmf over the word space $\{\beta_{ln}^{u_{ln}} \beta_{0n}^{1-u_{ln}}, n=1,\ldots,N\}$.
\end{enumerate}
\end{enumerate}

Note that $\mathbf{v}=\{v_{jd}\}$ indicate the topics present in each
document --
topic $j$'s probability ($\alpha_j v_{jd}$) of generating a word in document $d$ is non-zero only if $v_{jd}=1$.
Likewise, 
$(u_{jn}=1)$ if 
topic $j$ possesses a topic-specific probability $\beta_{jn}$ of generating word $n$;
otherwise, this probability is $\beta_{0n}$.
As aforementioned, the $\mathbf{u}$ and $\mathbf{v}$ switches together with the number of topics, $M$, specify the model \textit{structure} $\mathcal{H}(M,\mathbf{v},\mathbf{u})$. 
Given a fixed structure, the full complement of model {\it parameters} is given by $\Theta =\{\{\beta_{0n}\},\{\beta_{jn}\}, \{\alpha_{jd}\}\}$. 
Together, the structure $\mathcal{H}(M,\mathbf{v},\mathbf{u})$
 and parameters $\Theta$ constitute our model.

Given the above data generation mechanism, the likelihood of the corpus $\mathcal{D}$ under our model is:\vspace{-.1in}
\begin{flalign}
p(\mathcal{D|H},\hat{\Theta})=\prod_{d=1}^{D}\prod_{i=1}^{L_d}\sum_{j=1}^{M}\big[\alpha_{jd}v_{jd}\cdot{\beta^{u_{j{w_{id}}}}_{jw_{id}}}{\beta^{1-u_{j{w_{id}}}}_{0w_{id}}}\big].
\label{parsmdl}
\end{flalign}\vspace{0in}
Here, the double product appears because documents and words are generated i.i.d.
Also, we emphasize for clarity that $u_{j{w_{id}}}$ is topic $j$'s binary switch variable on word $w_{id}$,
the $i$-th word from document $d$.

The set of word probability parameters (both topic-specific and shared)
must be jointly constrained so that they give a valid pmf, conditioned on each topic, i.e. $(\sum_{n=1}^{N}{(u_{jn} \beta_{jn}+(1-u_{jn}) \beta_{0n})}=1)$, $\forall j$. Likewise, the topic proportions must satisfy the pmf constraints $\sum_{j=1}^{M}{\alpha_{jd}v_{jd}}=1$, $\forall d$. 
In the parameter learning step of our algorithm we estimate $\Theta =\{\{\beta_{0n}\},\{\beta_{jn}\}, \{\alpha_{jd}\}\}$ consistent with these constraints, assuming the model structure $\mathcal{H}(M,\mathbf{v},\mathbf{u})$ is known.
\vspace{-.15in}
\subsection{Parameter Estimation}
\label{ParamUpdate}
Based on the derivation given in the sequel in section \ref{bicder}, we seek the model which minimizes an objective function of the 
BIC form: (penalty term) $-$ (data log-likelihood).  Moreover, as will be seen, the penalty term 
in fact has {\it no} dependence on the model parameters, $\Theta$.  Thus, minimizing BIC with respect to $\Theta$ is
equivalent to maximizing the log-likelihood with respect to $\Theta$.  Accordingly, in our framework, 
the parameters $\Theta$ are chosen to (locally) maximize the log-likelihood of the data $\mathcal{D}$ via the Expectation Maximization (EM) algorithm \cite{Dempster1977}, as we next formulate. 

Let $Z_{id}$ be an M-dimensional binary random vector that has only a single element equal to one and all other elements equal to zero. Here, the non-zero element in $Z_{id}$ (i.e. $j$ s.t. $Z_{id}^{(j)}=1$) is the topic of origin for word $w_{id}$. We treat $Z=\{Z_{id}^{(j)}\}$ as the {\it hidden data} within our EM framework \cite{Dempster1977}.  At each iteration, EM maximizes a lower bound on the (incomplete) data log-likelihood,
with guaranteed monotonic increase in the incomplete data log-likelihood and locally optimal convergence \cite{Dempster1977}.
Each such iteration consists of i) an E-step, computing the expected value of the hidden data, given the observed data
and current parameter estimates, which is used to determine the expected {\it complete} data log-likelihood;
ii) An M-step, in which we update the model parameters by maximizing the expected complete data log-likelihood.
For our likelihood model (\ref{parsmdl}) and choice of hidden data, these two steps are specified as follows:

\noindent
{\it E-step:} 
Given the model structure and the current estimate of the parameters $\Theta^{(t)}$, we compute
\begin{flalign}
&E[Z_{id}^{(j)} | \mathcal{D};\Theta^{(t)},\mathcal{H}] = P(Z_{id}^{(j)}=1|w_{id};\Theta^{(t)},\mathcal{H})&\nonumber\\
&~~~~~~~~~~~~~~~=\frac{\alpha_{jd}v_{jd}{\beta^{u_{j{w_{id}}}}_{jw_{id}}}{\beta^{1-u_{j{w_{id}}}}_{0w_{id}}}}{\sum_{l=1}^{M}{\alpha_{ld}v_{ld}{\beta^{u_{l{w_{id}}}}_{lw_{id}}}{\beta^{1-u_{l{w_{id}}}}_{0w_{id}}}}},~~\forall i,d,
\label{pz}
\end{flalign}
i.e., the expected hidden data in this case are the posterior probabilities on topic origin $\forall id$.

Adding normalization constraints to the expected value of the \textit{complete} data log-likelihood measured with respect to (\ref{pz}), we construct the Lagrangian at the current parameter set estimate $\Theta^{(t)}$:\vspace{-0.05in}
\begin{align}
&E\big[\log(p(\mathcal{D},Z|\Theta,\mathcal{H}))|\Theta^{(t)},\mathcal{H}\big]=\sum_{d=1}^{D}\sum_{i=1}^{L_d}\sum_{j=1}^{M}\Big[v_{{jd}}\nonumber\\
&\cdot P(Z_{{id}}^{(j)}=1|w_{{id}};\Theta^{(t)},\mathcal{H})\Big(\log(\alpha_{{jd}})+u_{{j{w_{id}}}}\log(\beta_{jw_{{id}}}) \nonumber\\
&+(1-u_{{j{w_{id}}}})\log(\beta_{0w_{{id}}})\Big)\Big]-\sum_{d=1}^{D}{\lambda_d(\sum_{j=1}^{M}{\alpha_{{jd}}v_{{jd}}}-1)} \nonumber\\
&-\sum_{j=1}^{M}{\mu_{j}\Big(\sum_{n=1}^{N}{(u_{{jn}}\beta_{jn}+(1-u_{{jn}})\beta_{0n})}-1\Big)}.
\label{q}
\end{align}

\noindent
{\it M-step:}

Maximization with respect to topic proportions is achieved by setting the partial derivative of (\ref{q}) with respect to $\alpha_{jd}$ equal to zero and satisfying the normalization constraint:
\begin{flalign}
&\alpha_{jd}=\frac{\sum_{i=1}^{L_d}{P(Z_{id}^{(j)}=1|w_{id};\Theta^{(t)},\mathcal{H})v_{jd}}}{\sum_{l=1}^{M}{\sum_{i=1}^{L_d}{P(Z_{id}^{(l)}=1|w_{id};\Theta^{(t)},\mathcal{H})v_{ld}}}},&\nonumber \\
&~~~~~~~~~~~~~~~~~~~~~~~~~~~~~~~~~~~~~~~~\forall j,d, : v_{jd}=1.
\label{alphaupdate}
\end{flalign}

Similarly, maximization with respect to $\beta_{jn}$ is achieved by:\vspace{-0.05in}
\begin{equation}
\beta_{jn}=\frac{x_{jn}u_{jn}}{\mu_j},
\label{pj}
\end{equation}
\vspace{-0.2in}where\vspace{0.1in}
\begin{equation}
x_{jn}\triangleq \sum_{d=1}^{D}{\sum_{{i=1:{w_{id}=n}}}^{L_d}{P(Z_{id}^{(j)}=1|w_{id};\Theta^{(t)},\mathcal{H})v_{jd}}},\forall j,n. \nonumber
\end{equation}\vspace{-0.1in}

Here, $\mu_j$ is the Lagrange multiplier, which we solve for by multiplying both sides of (\ref{pj}) by $u_{jn}$, summing over all $n$, and applying the distribution constraint on topic $j$'s word model to obtain: \vspace{-0.1in}
\begin{equation}
\mu_{j}=\frac{\bar{x}_j}{1-\sum_{n=1}^{N}{(1-u_{jn})\beta_{0n}}},\bar{x}_j\triangleq \sum_{n=1}^{N}{x_{jn}u_{jn}},\forall j. 
\label{muj}
\end{equation}\vspace{-0.05in}

The E and M-steps are alternately applied, starting from initial parameter estimates, until a convergence
condition (assessing diminishing improvement in log-likelihood from one iteration to the next) is met.

\vspace{.05in}
\noindent
{\it Estimating the Shared Model Parameters:}

\vspace{.05in}
In principle, we can take the derivative of (\ref{q}) with respect to $\left\{\beta_{0n}\right\}$ and optimize it along with all other parameters. Alternatively, we can treat the shared model as a ``universal'' model that is estimated once, at initialization, and then held fixed while performing EM. Our experiments show that this latter approach is quite effective. Thus, in this work the shared model is estimated only once, via the global frequency counts:
\begin{equation}
\beta_{0n}=\frac{\sum_{d=1}^{D}{\sum_{i=1:{w_{id}=n}}^{L_d}{1}}}{\sum_{d=1}^{D}{L_d}},~~\forall n=1,...,N.
\label{p0}
\end{equation}

\section{BIC Derivation}
\label{bicder}
In this section we derive our BIC objective function, with respect to which we jointly optimize the entire model -- $\mathcal{H}(M,\mathbf{v},\mathbf{u})$ and $\Theta$. For fixed $M$ and $\Theta$, we will minimize BIC with respect to
$({\bf v}, {\bf u})$.
Alternately, given fixed structure, minimizing BIC with respect to $\Theta$ is achieved by EM as described in section \ref{ParamUpdate}.
Using such alternating minimization, locally BIC-optimal models are learned at each $M$. 
The BIC-minimizing model, at order 
$M^{\ast}$, is then chosen. 

The posterior probability of the model structure $\mathcal{H}$ is proportional to $p(\mathcal{D|H})p(\mathcal{H})$, $p(\mathcal{D|H})$ the marginal likelihood and $p(\mathcal{H})$, the structure prior. The marginal likelihood is computed by taking expectation of the data likelihood with respect to the prior distribution on the parameters:
\begin{equation}
I=p(\mathcal{D}|\mathcal{H})=\int{p(\mathcal{D}|\mathcal{H},\Theta)p(\Theta|\mathcal{H})d\Theta},
\label{bayesfactor}
\end{equation}
with $p(\Theta|\mathcal{H})$ the prior on parameters given structure $\mathcal{H}$. We use Laplace's method to approximate (\ref{bayesfactor}). This is based on the assumption that $p(\mathcal{D}|\mathcal{H},\Theta)p(\Theta|\mathcal{H})$ is peaked around its maximum (the posterior mode $\tilde{\Theta})$, which in fact is valid for large sample sizes \cite{Kass1995}. 

We approximate $\log\big(p(\mathcal{D}|\mathcal{H},\Theta)p(\Theta|\mathcal{H})\big)$ around the posterior mode using a Taylor series expansion. Note that the first Taylor term is constant with respect to the variables of integration and the term corresponding to first derivatives is zero. Thus, exponentiating the Taylor's series expansion at second order and plugging into (\ref{bayesfactor}), we obtain an approximation of the integral:\vspace{-.02in}
\begin{dmath}
\hat{I}=p(\mathcal{D}|\mathcal{H},\tilde{\Theta})p(\tilde{\Theta}|\mathcal{H})\int{e^{\big(-\frac{1}{2}(\Theta-\tilde{\Theta})^T\tilde{\Sigma}(\Theta-\tilde{\Theta})\big)}d\Theta},
\label{step0}
\end{dmath}\vspace{-.05in}
where $\tilde{\Sigma}$ is the negative of the Hessian matrix. 

The integrand in (\ref{step0}), the exponentiated second order Taylor series term, is a scaled normal distribution with mean $\tilde{\Theta}$ and covariance $\tilde{\Sigma}$, which, when integrated, evaluates to $(2\pi)^{k/2}{|\tilde{\Sigma}|}^{-1/2}$. Here $k$ is the number of parameters in $\Theta$, which is equal to $\sum_{d=1}^{D}{M_d}+\sum_{j=1}^{M}{N_j}$, where $M_d=\sum_{j=1}^{M}{v_{jd}}$ is the number of active topics in document $d$ and $N_j=\sum_{n=1}^{N}{u_{jn}}$ is the number of topic-specific words under topic $j$. Also, since the shared model parameters are estimated in a universal fashion, once, and then fixed (\ref{p0}), their descriptive complexity is assumed to be fixed, irrespective of $\mathcal{H}(M,\mathbf{v},\mathbf{u})$ and $\Theta$. Thus, this (assumed constant) term need not be included in our objective function.

We thus have the following approximation for the marginal likelihood \cite{Kass1995},\cite{Ghosh2006a}:\vspace{-.05in}
\begin{equation}
\hat{I}={(2\pi)}^{k/2}{|\tilde{\Sigma}|}^{-1/2}p(\mathcal{D}|\mathcal{H},\tilde{\Theta})p(\tilde{\Theta}|\mathcal{H}).
\label{laplace}
\end{equation}\vspace{-.15in}

Based on (\ref{laplace}), an approximate negative log-model posterior (BIC cost) is: \vspace{-.1in}
\begin{align}
&BIC=-\log(\hat{I})=-\frac{k}{2}\log{(2\pi)}+\frac{1}{2}\log{(|\tilde{\Sigma}|)}&\nonumber\\
&-\log(p(\mathcal{D}|\mathcal{H},\tilde{\Theta}))-\log(p(\tilde{\Theta}|\mathcal{H}))-\log(p(\mathcal{H}))\text{.}
\label{bic0}
\end{align}

As usually done, we take $p(\tilde{\Theta}|\mathcal{H})$ to be an uninformative prior, which can be neglected. In this way, we transition from a random description of the parameters $\Theta$ to treating them as deterministic unknowns.

By minimizing BIC, a tradeoff is achieved between the data log-likelihood, $\log(p(\mathcal{D}|\mathcal{H},\tilde{\Theta}))$ and the other terms, interpreted as the model ``complexity'' cost.

The negative of the Hessian matrix $\tilde{\Sigma}$, neglecting $p(\Theta|\mathcal{H})$, is:\vspace{-.1in}
\begin{equation}
[\tilde{\Sigma}]_{ij}=-\frac{\partial^2\log\big(p(\mathcal{D}|\mathcal{H},\Theta))}{\partial\Theta_i\partial\Theta_j}\bigg|_{\Theta=\tilde{\Theta}},i,j=1,2,...k.
\label{hessian}
\end{equation}\vspace{-.1in}

The diagonal elements corresponding to topic proportions are: \small
\begin{equation}
\hspace{-.1in}-\frac{\partial^2\log( p(\mathcal{D}|\mathcal{H},\Theta))}{\partial\alpha_{jd}^2}\Bigg|_{{\Theta=\tilde{\Theta}}}\hspace{-0.22in}=-v_{jd}\sum_{i=1}^{L_d}{\frac{\partial^2\log(p(w_{id}|\mathcal{H},\Theta))}{\partial\alpha_{jd}^2}}\Bigg|_{\Theta=\tilde{\Theta}}\hspace{-0.22in},\hspace{-0.22in}
\label{alphaderiv1}
\end{equation}\vspace{-.1in}
\normalsize
where
\begin{equation}
p(w_{id}|\mathcal{H},\Theta)=\sum_{l=1}^{M}{\alpha_{ld}v_{ld}{\beta^{u_{l{w_{id}}}}_{lw_{id}}}{\beta^{1-u_{l{w_{id}}}}_{0w_{id}}}}. 
\label{alphaderiv3}
\end{equation}

Invoking the weak law of large numbers, we have with probability 1 that
\begin{flalign}
-&\frac{\partial^2\log(p(\mathcal{D}|\mathcal{H},\Theta))}{\partial\alpha_{jd}^2}\bigg|_{\Theta=\tilde{\Theta}}& \\ &=-v_{jd}\frac{1}{L_d}\sum_{i=1}^{L_d}{L_d\frac{\partial^2\log(p(w_{id}|\mathcal{H},\Theta))}{\partial\alpha_{jd}^2}}\bigg|_{\Theta=\tilde{\Theta}} \nonumber \\ 
&\rightarrow-v_{jd}L_d{\frac{\partial^2E\big[\log(p(W_{id}|\mathcal{H},\Theta))\big]}{\partial\alpha_{jd}^2}\bigg|_{\Theta=\tilde{\Theta}}} = v_{jd}L_{d}I^{(\alpha)}_{jd},\nonumber
\label{alphaderiv2}
\end{flalign}
where 
$W_{id} \in \{1,\ldots,N\}$ is a random index and
$I^{(\alpha)}_{jd}$ is the corresponding element in the Fisher information matrix. Similarly, diagonal elements of the Hessian matrix corresponding to $\beta_{jn}$ are
\begin{align}
-&\frac{\partial^2\log(p(\mathcal{D}|\mathcal{H},\Theta))}{\partial{\beta^2_{jn}}}=& \\
&-u_{jn}\frac{1}{\bar{L}_j}\sum_{d=1}^{D}{\sum_{i=1}^{L_d}{\bar{L}_j\frac{\partial^2\log(p(w_{id}|H,\Theta))}{\partial{\beta^2_{jn}}}\Bigg|_{\Theta=\tilde{\Theta}}}}\nonumber \\
&\rightarrow-u_{jn}\big(\sum_{d=1}^{D}{\bar{L}_j}\big){\frac{\partial^2E\big[\log(p(W_{id}|H,\theta))\big]}{\partial{\beta^2_{jn}}}\Bigg|_{\Theta=\tilde{\Theta}}}\hspace{-0.22in}=u_{jn}{\bar{L}_j}I^{(\beta)}_{jn}\nonumber 
\label{betaderiv2}
\end{align}
Here $I^{(\beta)}_{jd}$ is the Fisher information matrix element corresponding to $\beta_{jn}$;  moreover, we identify $\bar{L}_j\equiv\sum_{d=1}^{D}{L_d{v_{jd}}}$.\vspace{.05in}

Following the common approach in deriving BIC \cite{Ghosh2006a}, we neglect the off-diagonal elements and write $\tilde{\Sigma}$ as a block-diagonal matrix,
\begin{equation}
\tilde{\Sigma}=
\begin{bmatrix}
\boldsymbol{\tilde{\Sigma}_{\alpha}} & \mathbf{0}\\
\mathbf{0} & \boldsymbol{\tilde{\Sigma}_{\beta}} \
\end{bmatrix}\text{, where } \begin{array}{l}\boldsymbol{\tilde{\Sigma}_{\alpha}}\triangleq Diag[L_dI^{(\alpha)}_{jd}],\\ \boldsymbol{\tilde{\Sigma}_{\beta}}\triangleq Diag[\bar{L}_jI^{(\beta)}_{jn}]. \end{array}
\label{sigmamatrix}
\end{equation}

As the sample size grows, the Fisher matrix terms in (\ref{sigmamatrix}) ($I^{(\beta)}_{jn}$ and $I^{(\alpha)}_{jd}$) become negligible \cite{Ghosh2006a}. Thus, we have:\vspace{-0.1in}
\begin{equation}
\log{(|\tilde{\Sigma}|)}\approx\sum_{d=1}^{D}{(M_d-1)\log(L_d)}+\sum_{j=1}^{M}{N_j\log(\bar{L}_j)}.
\label{samplesize}
\end{equation}\vspace{-0.1in}

This expression is reminiscent of the penalty term in the na\"{\i}ve BIC form where, for each parameter, one ``pays'' $\frac{1}{2}\log(sample~size)$. But here, in (\ref{samplesize}), we identify different penalties for each of the two different parameter types in our model, with the argument of the log() the {\it effective} sample size, consistent with our
BIC derivation. That is, this effective size is   
$L_d$ for an $\alpha_{jd}$ parameter and $\bar{L}_j$ for a $\beta_{jn}$ parameter.
However, we can also sanity-check this interpretation against our parameter re-estimation equations.  Note in particular
that the $\alpha_{jd}$ parameters are indeed re-estimated in (\ref{alphaupdate}) based on a sum over $L_d$ ``word samples'' and 
the $\beta_{jn}$ parameters are re-estimated in (\ref{pj}) based on a sum over $\bar{L}_j$ ``total word samples over all documents
in which topic $j$ is present''.  {\it Thus, indeed, our derivation leads to a generalization of the standard BIC cost, wherein the penalty on a parameter of given
type is} $\frac{1}{2}\log(\textit{effective~sample~size})$.

The other important term in (\ref{bic0}) is the prior probability of the model structure $p(\mathcal{H})$, defined based on the configurations of $\mathbf{u}$ and $\mathbf{v}$ switches for a given number of topics $M$. Here, we assume that priors on $\mathbf{u}$ and $\mathbf{v}$ switches are independent. 

Our general principle in defining these prior distributions is to invoke uninformativeness (uniformity). We parameterize the prior distribution on $\mathbf{u}$ switches as a function of the total number of topic-specific words across all topics, $N_0$, and (uninformatively) assume all configurations with at most $N_0$ ``on'' switches ($\sum_{n=1}^{N_0}{\binom{MN}{n}}$) are equally likely. However, under the assumption that, generally, a small number of words across all topics will be topic-specific ($N_0 \ll NM$), the total number of these configurations can be well-approximated by \cite{Graham1994}: \vspace{-0.08in}
\begin{equation}
\sum_{n=1}^{N_0}{\binom{MN}{n}}\approx \frac{2^{MNH(\frac{N_0}{MN})}}{\sqrt{MN}},
\label{ujnum}
\end{equation}
where $H(\cdot)$ is Shannon's entropy for a Bernoulli random variable with parameter $\frac{N_0}{MN}$ \cite{M.Cover2006}. Accordingly, the corresponding term in BIC is:\vspace{-0.08in}
\begin{equation}
-\log(p(\mathbf{u}))=MNH(\frac{\bar{N}}{N})\log(2)-\frac{1}{2}\log(MN),
\label{ucost}
\end{equation}
where $\bar{N}\triangleq\frac{1}{M}N_0 = \frac{1}{M}\sum_{j=1}^{M}{N_j}$ is the average number of topic-specific words across all topics.

To define the prior on $\mathbf{v}$ configurations, we consider a two stage process for each document, $d$. First, the number of topics ($M_d$) is selected from a uniform distribution ($\frac{1}{M}$); then a switch configuration is selected from a uniform distribution over all $\binom{M}{M_d}$ configurations with $M_d$ ``on'' switches. Therefore, \vspace{-0.08in}
\begin{equation}
-\log(p(\mathbf{v}))=D\log(M)+\sum_{d=1}^{D}{\log\binom{M}{M_d}}.
\label{vcost}
\end{equation}

By substituting (\ref{samplesize}), (\ref{ucost}), and (\ref{vcost}) into (\ref{bic0}), our overall BIC expression becomes:
\begin{flalign}
&BIC=D\log(M)+\sum_{d=1}^{D}{\log\binom{M}{M_d}}+MNH(\frac{\bar{N}}{N})\log(2)&\nonumber\\
&~~~~-\frac{1}{2}\log(MN)+\frac{1}{2}\sum_{d=1}^{D}{(M_d-1)\log(\frac{L_d}{2\pi})}\nonumber\\
&~~~~+\frac{1}{2}\sum_{j=1}^{M}{N_j\log(\frac{\bar{L}_j}{2\pi})}-\log(p(\mathcal{D}|\mathcal{H},\tilde{\Theta}))\text{~.\footnotemark}
\label{bic}
\end{flalign}
\footnotetext{We remind the reader that, though this explicit dependence is omitted, $N_j$ is a function of ${\bf u}$ and $M_d$ is a function of ${\bf v}$.}
\vspace{-.1in}
\section{Integrated Model Selection and Parameter Estimation}
\label{mainalg}
In this section we develop our algorithm for jointly determining $\mathcal{H}(M,\mathbf{v},\mathbf{u})$ and $\Theta$ to minimize the BIC objective (\ref{bic}).
Supposing, for the moment, that $M$ is fixed, there are $2^{MD+NM}$ different possible $(\mathbf{u},\mathbf{v})$ switch configurations -- global minimization of BIC over this search space is a formidable task. However, using a generalized Expectation Maximization (GEM) algorithm \cite{Dempster1977},\cite{Meng1997}, we can break up this problem into a series of simple update steps, each decreasing in BIC, with the overall algorithm thus guaranteed to converge to a local minimum.

To formulate this approach, we note that the EM framework is applicable to monotonically descend in the BIC cost (\ref{bic}), 
just as it is applied to monotonically ascend in the incomplete data log-likelihood $\log(p(\mathcal{D}|\mathcal{H},\tilde{\Theta}))$.
Toward this end, we define the {\it expected complete data} BIC cost.  This is formed by: 1) substituting in the 
complete data log-likelihood $\log(p(\mathcal{D},Z|\mathcal{H},\tilde{\Theta}))$, to replace the incomplete data log-likelihood $\log(p(\mathcal{D}|\mathcal{H},\tilde{\Theta}))$, in (\ref{bic}),
yielding the complete data BIC;
2) Taking the expected value of the complete data BIC.  The only term in the complete data BIC that is a function of the (random)
hidden variables is the complete data log-likelihood.  Thus, the expected value of the complete data BIC is formed simply by replacing the incomplete data log-likelihood term in (\ref{bic}) by the expected complete data log-likelihood term from (\ref{q}).

This expected complete data BIC is the keystone quantity for defining our GEM algorithm,
which descends in (\ref{bic}) while iteratively re-estimating the pair $(\mathcal{H},\Theta)$.  
Given a fixed model order $M$, our GEM algorithm consists of the following steps, iterated until convergence: 
\begin{enumerate}
\item\emph{E-step}: Following section \ref{ParamUpdate}, expected values of the hidden variables are given in (\ref{pz}). Based on 
the discussion above,
the expected complete data BIC is formed by replacing the incomplete data log-likelihood term in (\ref{bic}) by the expected complete data log-likelihood term from (\ref{q}).
\item\emph{Generalized M-step}:
\begin{enumerate}
\item Estimation of $\Theta$ given fixed structure: Since the model complexity terms in BIC have no dependence on $\Theta$, minimization of the expected complete data BIC with respect to $\Theta$ given fixed structure is equivalent to maximizing the expected complete data log-likelihood, with closed form updates as described in section \ref{ParamUpdate}. 
\item Optimization of $(\mathbf{u},\mathbf{v})$ given fixed $\Theta$.
\end{enumerate}
\end{enumerate}
	
Updating the $(\mathbf{u},\mathbf{v})$ switches given fixed parameters (step 2(b)) is done via an iterative loop in which switches are cyclically visited one by one, with all parameters and other switches fixed. If a change in the current switch reduces BIC, that change is accepted.\footnote{For computational efficiency, optimization over $\mathbf{u}$ is with respect to the expected complete data BIC. However for $\mathbf{v}$, we directly minimize BIC since there is no computational advantage in minimizing with respect to expected BIC in this case. Since minimizing expected complete data BIC ensures descent in BIC, both $\mathbf{u}$ and $\mathbf{v}$ update steps descend in the BIC objective (\ref{bic}).} This process is repeated over all switches until no further decrease in BIC occurs or a predefined maximum number of cycles is reached. We update the $\mathbf{u}$ and $\mathbf{v}$ switches separately, first performing cycles over all $\mathbf{u}$ switches until convergence and then over the $\mathbf{v}$ switches until convergence. We then go back to the E-step.

A change in one switch affects both the model cost ($\Delta Cost$) and log-likelihood ($\Delta L$) terms in BIC. In the following sections, we determine ($\Delta BIC=\Delta Cost-\Delta L$) associated with switch updates.

\subsection{Updating \{$u_{jn}$\}}
\label{uupdate}
Here, we trial-flip each switch $u_{j'n'}, n'=1,...,N$ in every topic $j'=1,...,M$, one by one. In performing these updates, each topic is constrained to have at least one topic-specific word. 

In the model cost terms of BIC, a change in the value of $u_{j'n'}$ only affects the number of topic-specific words in that topic, $N_j'$. Thus,
\begin{flalign}
&\Delta Cost(u_{j'n'})=\frac{1}{2}{(-1)}^{u^-_{j'n'}}\log(\frac{\bar{L}_{j'}}{2\pi})&\nonumber\\
&~~~~~~~~~~~~~~~~~+NM\log(2)\big(H(\frac{\bar{N}^+}{N})-H(\frac{\bar{N}^-}{N})\big),\hspace{-0.1in}
\label{dcostu}
\end{flalign}
where $^-$ and $^+$ superscripts denote the current and new values of functions of the switch, respectively.

In computing the change in the log-likelihood, we need the corresponding word probability under the new and old  values of the switch. Introducing the variables $x_{jn}$ and $\bar{x}_j$ from (\ref{pj}) and (\ref{muj}) allows us to 
compute $\mu^+_{j'}$ efficiently
(without performing the E-step and recomputing the required statistics). The effect of changing $u_{j'n'}$ on the expected value of the complete data log-likelihood is:\vspace{-0.1in}
\begin{equation}
\Delta L(u_{j^{'}n^{'}})= {(-1)}^{{u^-_{j'n'}}}x_{j'n'}\log(\frac{x_{j'n'}}{\mu^+_{j'}\beta_{0n'}})-\bar{x}^-_{j'}\log(\frac{\mu^+_{j'}}{\mu^-_{j'}}),
\label{dqu}
\end{equation}\vspace{-0.1in}
where:
\begin{equation}
\mu^+_{j'}=\frac{\bar{x}^-_{j'}+{(-1)}^{u^-_{j'n'}}x_{j'n'}}{1-\sum_{n=1}^{N}{(1-u^-_{j'n})\beta_{0n}}+{(-1)}^{u^-_{j'n'}}\beta_{0n'}}.
\label{munew}
\end{equation}

\subsection{Updating \{$v_{jd}$\}}
\label{vupdate}
We update the $\mathbf{v}$ switches by sequentially visiting each topic $v_{j'd'}~ j'=1,...,M$ in every document $d'=1,...,D$. Obviously, each document should be constrained to have at least one active topic. Moreover, every topic is constrained to be used by at least one document. 

Trial-flipping switch $v_{j'd'}$ changes $M_{d'}$ and $\bar{L}_{j'}$. Therefore,\vspace{-0.1in}
\begin{flalign}
&\Delta Cost(v_{j'd'})=\log(\frac{\binom{M}{M_{d'}^+}}{\binom{M}{M_{d'}^-}})+\frac{1}{2}\cdot\bigg[{(-1)}^{v^-_{j'd'}}\log(\frac{L_{d'}}{2\pi})&\nonumber\\
&~~~~~~~~~~~~~~~+N_{j'}\log{\Big(\frac{\bar{L}^-_{j'}+{(-1)}^{v^-_{j'd'}}L_{d'}}{\bar{L}^-_{j'}}\Big)}\bigg].
\label{dcostv}
\end{flalign}

Unlike the $\mathbf{u}$ updates, since expected values of hidden variables are zero for topics $j'$ such that $v_{j'd'}=0$, we need to recompute (\ref{pz}) for document $d'$ and trial-update $\alpha_{jd'}$ $\forall~j=1,..,M$ using (\ref{alphaupdate}). Then, we compute $\Delta L(v_{j'd'})$ based on the incomplete data log-likelihood: 
\begin{equation}
\Delta L(v_{j'd'})=\sum_{i=1}^{L_{d'}}{\log{\Big(\frac{\sum_{j=1}^{M}{\alpha^{+}_{jd'}v^{+}_{jd'} {\beta^{u_{jw_{id'}}}_{jw_{id'}}}{\beta^{1-u_{jw_{id'}}}_{0w_{id'}}}}}{\sum_{j=1}^{M}{\alpha^{-}_{jd'}v^{-}_{jd'} {\beta^{u_{jw_{id'}}}_{jw_{id'}}}{\beta^{1-u_{jw_{id'}}}_{0w_{id'}}}}}\Big)}}.
\label{dqv}
\end{equation}

Note that since all topic proportions in document $d'$ change when flipping one switch, unlike computing $\Delta L(u_{j^{'}n^{'}})$, there is no computational benefit in using the expected value of the complete data log-likelihood. This is why we are evaluating the incomplete data log-likelihood for these updates. 

\subsection{Initialization}
It is important to initialize the model sensibly to avoid poor local minima. Here, we use a simple but pragmatic initialization process:
 1) to initialize each topic, we randomly select $D_{init}$ documents. Only the words that occur in these documents are initially chosen as topic-specific, and their probabilities are initialized via frequency counts; 2) 
Based on this initial model, we use a maximum likelihood decision rule to hard-assign each document to a single topic;
3) Using this now more ``refined'' set of documents for each topic, we re-estimate topic-specific word probabilities via frequency counts. 

\subsection{Computational Complexity}
For fixed $M$, the computational complexity of our parameter learning is $O(M_DDL_D)$ where $M_D$ and $L_D$ are the number of topics present in a document and the length of a document, respectively. Since in LDA all topics are potentially present in each document, its computational complexity is $O(MDL_D)$, higher than for our model. However, structure learning in our model imposes further complexity. 

Updates of word switches involve an iterative loop over all words in all topics. But the trial-flip of a single switch only requires scalar computation. Thus, complexity of this part of structure learning is $O(MN)$. 

Experimentally, we find that the heaviest part of our algorithm is updating the $\mathbf{v}$ switches, where at each step we trial-estimate topic proportions and evaluate the incomplete data log-likelihood for the current document under consideration. We need computations of order $O(M_DL_D)$ for trial-update of each switch. Thus the total complexity for this part, including the loop over all $\mathbf{v}$ switches, is $O(MM_DDL_D)$. 

Overall, computational complexity of our model, $O(MM_DDL_D+MN)$, is higher than for LDA. Complexity may be an issue for a corpus with a large number of active topics across documents. However, we have also investigated complexity experimentally, via recorded execution times, and have found that our method and LDA require comparable execution. (cf. Table \ref{runtime}). 

\subsection{Model Order Selection}
We compare estimated models at different orders and choose the one with minimum BIC. Different strategies are conceivable for learning models over a range of orders. 
A sensible approach is to learn a model at some order and use it for initializing new models at other orders. In this way, we can sweep a range of model orders in either a bottom-up or top-down fashion. 
In the top-down approach, we initialize our model with a specified ``ceiling'' number of topics $M_{max}$ and reduce the number of topics by a predefined step.  We remove the least ``plausible'' topics from the existing model and apply the learning algorithm to minimize BIC at the now reduced model order, using the current set of model parameters as initialization. Here, as ``least plausible'', we simply remove the topics with the smallest aggregate mass across the document corpus. This is applied repeatedly until a minimum number of topics is reached. Experimentally, this method has been found to be superior to an alternative ``bottom-up'' approach. Thus, we have applied
the top-down method in our experiments.

\section{Experimental Results}
\label{results}
In this section we report results of our model on the Ohsumed, Reuters-21578, and 20-Newsgroup corpora as well as a subset of the LabelMe \cite{russell2008labelme} image dataset. For each dataset, we compared against LDA\footnote{http://www.cs.princeton.edu/${\sim}$blei/lda-c/} with respect to  model fit (training and held-out log-likelihood), sparsity, and a
class label consistency measure. We also compared against an extension of LDA (LDAbackground) which includes one additional topic (a background topic) whose probabilities are fixed at the global frequency estimates. Furthermore, we compared against STC\footnote{http://www.ml-thu.net/~jun/stc.shtml}\cite{Zhu2011} with respect to class label consistency\footnote{Since STC is not a generative modelling approach, it is not suitable to evaluate it with respect to data likelihood.}. C implementation of our model is available from {https://github.com/hsoleimani/PTM}. 

\cite{Hoffman2012} suggests that comparing log-likelihoods computed by different topic models may be unfair, since each may compute a different approximation of the log-likelihood. Here, we use the method described in \cite{Hoffman2012}, \cite{Teh2007} to compare model fitness on a held-out test set. In this approach, we divide each document in the test set into two sets, observed and held-out parts, keeping the set of unique words in these two parts disjoint. We use the observed part of the test set to compute the expected topic proportions and then compute the log-likelihood on the words in the held-out set:
\begin{equation}
\sum_{d=1}^{D_{test}}\sum_{i=1}^{L^{heldout}_d}\log\Big(\sum_{j=1}^{M}E_q\big[\alpha_{jd}\big]E_q\big[\beta_{jw_{id}}\big]\Big).
\end{equation}

Note that for LDA, $E_q\big[\alpha_{d}\big]$ are the variational parameters $\gamma_d$ normalized to sum to one, while for our model they are directly the topic proportions $\alpha_d$. $E_q\big[\beta_{jn}\big]$ is $u_{jn}\beta_{jn}+(1-u_{jn})\beta_{0n}$ in our model. In LDA, however, $E_q\big[\beta_{jn}\big]$ is simply equal to the corresponding estimated word probability.

Topic models can also be compared with respect to ``quality'' of extracted topics. From a computational linguistics standpoint, topics are expected to exhibit a coherent semantic theme rather than covering loosely related concepts. In this paper, we use the criterion proposed in \cite{Mimno2011} to evaluate \textit{coherence} of learned topics. This measure has been shown to be in agreement with experts' coherence judgments \cite{Mimno2011}. The \textit{topic coherence} of topic $j$, which is based on word co-occurrence frequencies, is computed as:
\begin{equation}
C(j;N^{(j)})=\sum_{k=2}^{T}{\sum_{l=1}^{k-1}{\log{\frac{S(n_k^{(j)},n_l^{(j)})+1}{S(n_l^{(j)})}}}},
\label{coh}
\end{equation}
where $N^{(j)}=(n_1^{(j)},...,n_T^{(j)})$ are the $T$ most probable topic-specific words under topic $j$ and $S(n,n')$ is the number of documents in the corpus containing both words $n$ and $n'$. Similarly, $S(n)$ represents the number of documents with word $n$. Here, we use the top 2 percent of topic-specific words in each topic for our model to compute coherence.  We use the same number of top words to compute coherence for LDA topics. In our experiments, we report the average coherence over all topics.

Results reported here are averaged based on 10 different initializations. For STC, we determined the hyper-parameters by a validation set approach, working on the model with highest number of topics, and then kept them fixed for all other model orders. This makes STC complexity manageable and is also reasonable because we observe that the best STC accuracies are anyway achieved at the highest model order. The validation set was created by randomly choosing 20\% of the documents in the training set. For inference on test documents, for our model, we allow all topics to be active in each document and only optimize topic proportions using (\ref{alphaupdate}), given a fixed number of topics. We have taken this approach rather than using a transductive inference approach. 

\subsection{Ohsumed Corpus}
Ohsumed\footnote{http://disi.unitn.it/moschitti/corpora.htm} consists of 34389 documents, each assigned to one or multiple labels of the 23 MeSH diseases categories. Documents have on average $2.26$ (std. dev. = $0.84$) labels. The dataset was randomly divided into 24218 training and 10171 test documents. There are 12072 unique words in the corpus after applying standard stopword removal.

\begin{figure}
\centering
\includegraphics[scale=1]{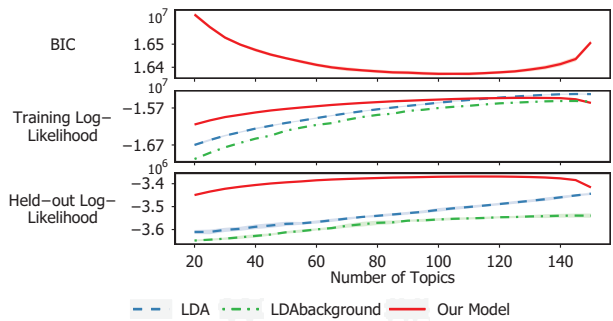}
\vspace{-0.05in}
\caption{Average performance, based on 10 random initialization, for the Ohsumed corpus.}
\vspace{-.2in}
\label{figumed}
\end{figure}

For all four methods, models were initialized with $150$ topics and at each step the five least massive topics were removed. Fig. \ref{figumed} shows the BIC curve and the training and held-out log-likelihood of our model compared to LDA and LDAbackground. The minimum of the BIC curve (i.e. estimated number of topics) is on average $M^*=105$ (std. dev. = $5.98$). The figure shows that LDA achieves higher log-likelihood on the training set for models with more than $120$ topics; but by controlling the likelihood-complexity trade-off, our model achieves higher log-likelihood on the held-out set at all orders.  

We also used the class labels provided with the dataset to evaluate a class label consistency measure. We first associated to each topic a multinomial distribution on the class labels. We learned these label distributions for each topic by frequency counting over the ground-truth class labels of all documents, weighted by topic proportions: 
\begin{align}
p_j(c)=\frac{\sum_{d=1}^{D}{\sum_{i=1:l_{id}=c}^{|C_d|}{\alpha_{jd}v_{jd}}}}{\sum_{d=1}^{D}{\sum_{i=1}^{|C_d|}{\alpha_{jd}v_{jd}}}},~\forall j,c,
\label{labeling_learn}
\end{align}
where $l_{id}$ is the $i$-$th$ class label in document $d$ and $|C_d|$ and $|C|$ are, respectively, the number of labels for document $d$ and the total number of class labels. For labeling a test document, we then compute the probability of each class label based on the topic proportions in that document; i.e. $\sum_{j=1}^{M}{\alpha_{jd}p_j(c)}~ c=1,...,|C|$, and assign the labels that have probability higher than a threshold value $\nu$:\vspace{-0.07in}
\begin{equation}
\hat{C}_d = \bigg\{c:\sum_{j=1}^{M}{\alpha_{jd}p_j(c)}>\nu\bigg\}
\label{labeling_assign}
\end{equation}\vspace{-0.08in}

As label consistency criteria, we measure \textit{precision} and \textit{recall} on the test set. Precision is the number of true discovered labels divided by the total number of ground-truth labels. Recall is the number of correctly classified labels divided by the total number of labels assigned to documents by our classifier. We measure these criteria for different threshold values $\nu$ and report the area under the precision/recall curve (AUC) as the final measure of performance. For LDA and LDAbackground we used the normalized Dirichlet variational parameters $\gamma_j^{(d)}$ in (\ref{varparams}) as topic $j$'s proportion for document $d$. Similarly, we used STC's normalized document codes for its unsupervised classification. For both LDA and STC we set $v_{jd}=1~\forall j,d$.

\begin{figure}[t]
\includegraphics[scale=1]{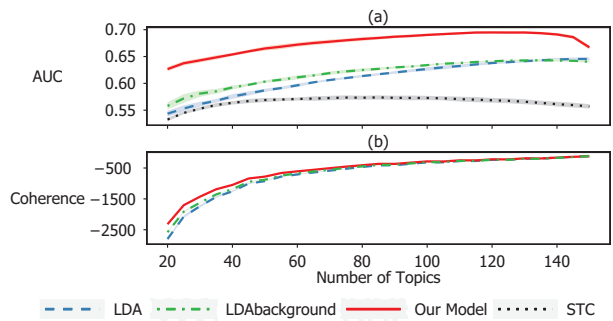}
\vspace{-0.05in}
\caption{(a) Class label consistency and (b) coherence versus number of topics on Ohsumed.}
\label{c}
\end{figure}

Fig. \ref{c}a shows the AUC for our model, LDA, LDAbackground and STC. We see that adding a background model to LDA improves class label consistency. However, the best performance of our model is better than other methods' over the entire range of model orders. Also, the highest AUC for our model occurs near $M^*=105$, which is the minimum of the BIC curve. 

Average coherence over all topics for our model, LDAbackground and LDA are plotted in Fig. \ref{c}b. This figure shows that the average coherence in our model is higher than LDA and LDAbackground for a wide range of model orders.

Tables \ref{mbar_tab} and \ref{nbar_tab} compares sparsity in our model against LDA, LDAbackground, and STC at $M=105$. As the measure of topic proportion sparsity, we report the average number of occurring topics in documents ($\overline{M}$). In STC and LDA, we considered those topics which dominantly contributed in generating at least one word in a document as ``occurring topics''. We also report average number of topic-specific words per topic ($\overline{N}$), and the total number of topic-specific words over all topics ($N_{unique}$) for our model. 

Average number of occurring topics in our model is $1.85$ which compared to LDA and STC is much closer to the average number of labels per document in this corpus ($2.26$). This suggests our topics better resemble the ground-truth classes than those of LDA and STC. Also, the shared model is widely used in our model and only a relatively small number of words are specific for each topic. Average number of unique topic-specific words over all topics is much larger than $\overline{N}$, which indicates that there is not great overlap between the set of topic-specific words across topics. In fact, topic-specific words, on average, are salient in $8.10$ topics (out of $105$) and modeled by the shared representation in other topics.
\begin{table*}
\caption{Sparsity on topic proportions; $\overline{M}:$ Average number of topics present in a document;(std. dev.)}
\centering
\begin{tabular}{ |c|c|c|c|c| }
\hline 
Dataset & Ohsumed ($M$=105)& 20-Newsgroup ($M$=36)& Reuters ($M$=40)& LabelMe ($M$=46)\\ \hline
Our Model	&	1.85 (0.01)	&	1.14 (0.01)	&	1.17 (0.01) & 7.4 (0.14)\\ \hline
LDA &	10.55 (0.37)	&	6.42 (0.28) &	4.97 (0.13) & 37.15 (1.03)\\ \hline
LDAbackground &	12.49 (0.31)	&	5.27 (0.12) &	4.18 (0.74) & 36.17 (0.71)\\ \hline
STC	&	10.22 (0.50)	&	2.67 (0.06) &	3.01 (0.20) & 1.08 (0.04)\\ \hline
\end{tabular}
\label{mbar_tab}
\vspace{-.1in}
\end{table*}
\begin{table}
\caption{Sparsity measure on topic-specific words for our model; $\overline{N}:$ Average number of topic-specific words per topic; $N_{unique}:$ Average number of unique topic-specific words over all topics; (std. dev.)}
\centering
\begin{tabular}{ |c|c|c|c| }
\hline 
~ & $N$  & $\overline{N}$ & $N_{unique}$ \\ \hline
Ohsumed	&	12072	&863.04	(3.20)	&	11182 (11.96)\\ \hline
20-Newsgroup &54520 &	1640 (20.00) & 22070 (352) \\ \hline
Reuters &	16000	&506.4 (4.51)	&7149.8 (55.24) \\ \hline
LabelMe & 158 & 95.73 (2.31) & 158 (0) \\ \hline
\end{tabular}
\label{nbar_tab}
\vspace{-.1in}
\end{table}

\subsection{20-Newsgroups Corpus}
In this section we report the results of our comparison on 20-Newsgroups\footnote{http://people.csail.mit.edu/jrennie/20Newsgroups/}. This dataset consists of, respectively, 11293 and 7527 documents in the training and test sets, with 20 designated newsgroups. Each document is labeled to a single newsgroup.  After standard stopword removal and stemming there are 54520 unique words in the corpus. 

\begin{figure}[t]
\centering
\includegraphics[scale=1]{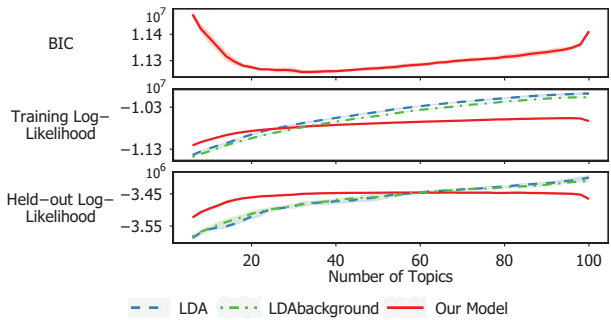}
\vspace{-0.05in}
\caption{Average performance, based on 10 random initializations, versus number of topics on 20-Newsgroups.}
\vspace{-.1in}
\label{fig20ng}
\end{figure}

Models were initialized with $100$ topics and two topics were removed at each elimination step. Fig. \ref{fig20ng} shows the performance of our model relative to LDA and LDAbackground. The minimum of the BIC curve is on average $M^*=36$ (std. dev. = $3.79$). Although LDA achieves higher likelihood on the training data, our parsimonious model has better performance on the held-out set for models with less than 60 topics. 

\begin{figure}
\includegraphics[scale=1]{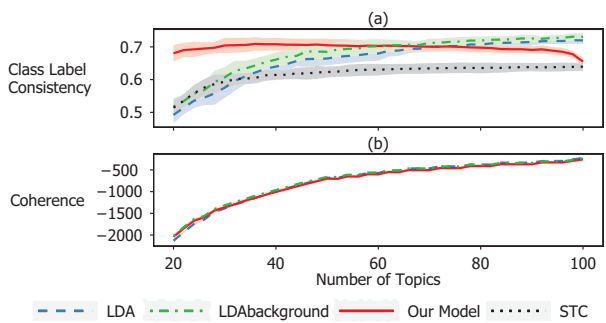}
\vspace{-0.05in}
\caption{(a) Class label consistency and (b) coherence versus number of topics on 20-Newsgroups.}
\vspace{-.15in}
\label{b}
\end{figure}

Learning the label distributions of topics is done similar to the procedure explained for the Ohsumed corpus. But since documents in this corpus have single labels, we compute the probability over the class labels for each document, $\sum_{j=1}^{M}{\alpha_{jd}p_j(c)}~ c=1,...,|C|$, and assign only the label with highest probability. Class consistency on the test set is reported in Fig. \ref{b}a. We can see that the BIC-chosen model order $M^*=36$ achieves good class consistency. Also, our model achieves better consistency compared 
to all other methods for models with less than $70$ topics. 
Coherence of the discovered topics in LDA, LDAbackground and our model is plotted in Fig. \ref{b}b. The curves are very similar.

Sparsity measures of our model on this dataset are compared with LDA, LDAbackground and STC in Tables \ref{mbar_tab} and \ref{nbar_tab}. Only a small fraction of the unique words are used as topic-specific words in our model. We can see that although adding a background model reduces the average number of occurring topics in LDA, our model still has the smallest $\overline{M}$ ($1.14$) among all methods. Note again that documents in this corpus are single-labeled. On average, the $22070$ unique topic-specific words in this dataset are context-specific in only $2.67$ topics, and shared in the others.

The 10 words with highest probabilities from some of the topics discovered by LDA and by our model are reported in Table \ref{sample_topics}. For our model, we separately report top 10 topic-specific and shared words. Note that some high probability words in LDA topics (``write'' and ``articl'' in the first, and ``system'' in the third topic), are shared words under our model.
\begin{table*}[ht]
\scriptsize
\caption{Top 10 words for sample topics extracted from the 20-Newsgroup and the Reuters corpora by our method and LDA. ``Specific'' and ``shared'' denote, respectively, the top topic-specific and shared words of the topics in our model.}
\centering
\begin{tabular}{ l||p{7.5cm}||p{7.5cm} }
\hline 
Model & {20-Newsgroup Topics} & {Reuters Topics} \\\hline
LDA & medic diseas articl write patient pitt bank health gordon doctor & gold mine ounc dixon miner ton dome south year feet\\
Shared & write articl don time work year good even problem thing & compani oper corp thi two unit includ expect report plan\\
Specific & doctor patient diseas medic pain peopl pitt treatment gordon bank & gold mine ounc ton year feet pct reserv mln dlr \\\hline
LDA & god christian jesu israel church peopl bibl christ come law& oil opec ga price energi dlr barrel mln petroleum crude \\
Shared & write articl time good even apr come find sinc gener & stock last two march industri expect quarter per month report\\
Specific & god christian exist peopl don atheist question thing mean reason& oil opec price bpd mln saudi barrel crude dlr product\\ \hline
LDA & drive problem disk window system work hard run driver printer& dividend april record split stock march declar payabl set share\\
Shared & write work good system want apr two question post program & corp offer unit quarter plan exchang invest acquir propos pai\\
Specific & printer font print window problem driver deskjet laser file articl & dividend stock share split april record compani common declar sharehold\\ \hline
\end{tabular}
\vspace{-.1in}
\label{sample_topics}
\end{table*}

\subsection{Reuters Corpus}
We compared our model against LDA, LDAbackground and STC on a subset of the Reuters dataset-21578\footnote{http://www.daviddlewis.com/resources/testcollections/reuters21578} consisting of documents from 35 classes. There are respectively 6454 and 2513 documents in the training and test sets, each labeled with a single class. The documents include 16000 unique words after applying standard stopword removal and stemming. 

\begin{figure}
\centering
\includegraphics[scale=1]{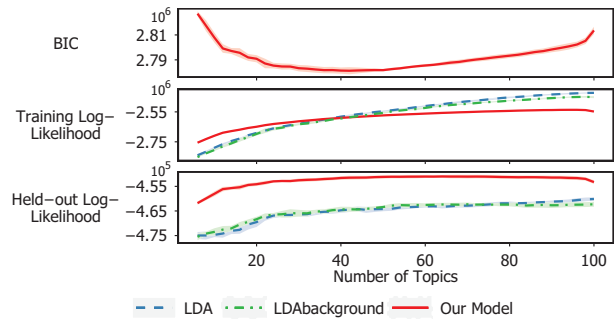}
\vspace{-0.05in}
\caption{Average performance, based on 10 random initializations, versus number of topics on Reuters.}
\vspace{-.1in}
\label{figr52}
\end{figure}

We initialized the models with $100$ topics and removed two topics at each elimination step. Average BIC for our model as well as the training and held-out data log-likelihood of our model, LDA and LDAbackground are shown in Fig. \ref{figr52}. The minimum of the BIC curve is on average at $M^*=40$ (std. dev. = $5.75$). Fig. \ref{figr52} shows that our model at $M=40$ achieves higher log-likelihood on the held-out set compared to the LDA models at all orders.  Moreover, this in spite of the fact that the held-out likelihood for
our model increases modestly for orders above $M=40$. 

\begin{figure}
\includegraphics[scale=1]{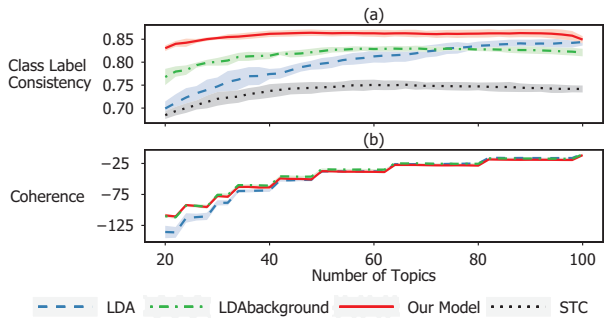}
\vspace{-0.05in}
\caption{(a) Class label consistency and (b) coherence versus number of topics on Reuters.}
\vspace{-.1in}
\label{cc}
\end{figure}

Class label consistency on this dataset is evaluated based on the procedure described for the 20-Newsgroup corpus. Average test set classification accuracy of our model, LDA, LDAbackground, and STC for this dataset are shown in Fig. \ref{cc}a. Our model has best performance across all model orders. Also, the minimum of the BIC curve, $M^*=40$, is consistent with high class label consistency. 

We plotted the average coherence of topics for our model and LDA in Fig. \ref{cc}b. This figure shows that for model orders less than $50$, topics discovered by our model have on average higher coherence than LDA. 

Tables \ref{mbar_tab} and \ref{nbar_tab} report the sparsity measures for our model, LDA, LDAbackground, and STC. The topic-specific words in our model are a small subset of the dictionary size. Also, our model achieves sparser topic presence compared to LDA, LDAbackground, and STC.

Table \ref{sample_topics} shows the top 10 words from three sample topics extracted by our model and LDA. For our model, we separately report the top 10 topic-specific and shared words. 

\subsection{LabelMe Image Datatset}
In this section we report the results of our comparison on a subset of the LabelMe image dataset. The dataset which we downloaded\footnote{http://www.cs.cmu.edu/${\sim}$chongw/slda/} consists of 1600 images from 8 classes. Unique codewords were extracted from the images using the method described in \cite{Fei-Fei2005}, \cite{Blei2009}. First, 128-dimensional SIFT vector descriptors \cite{Lowe1999} were generated by $5\times 5$ sliding grids for each image in the training set. Then, K-means clustering was performed on the collection of SIFT descriptors, giving learned cluster centers. Finally, each SIFT descriptor in every image was assigned to its nearest cluster \cite{Leung2001}. There are 158 clusters in this dataset after performing K-means clustering, merging close clusters, and pruning clusters with small number of members. The cluster index set $\left\{1,2,...,158\right\}$ is effectively the dictionary of words. Each image is represented by a sequence of these cluster indices, of length 2401.

Since the number of codewords in this dataset is very small relative to the text corpora, the approximation used in (\ref{ujnum}) is not valid anymore. Accordingly, we considered any configuration with at most ($N_0=NM$) ``on'' switches, equally likely and used the \textit{exact} form of (\ref{ujnum}). Thus, the cost of topic-specific words in BIC is $NM\log(2)$ in this case. 

\begin{figure}
\includegraphics[scale=1]{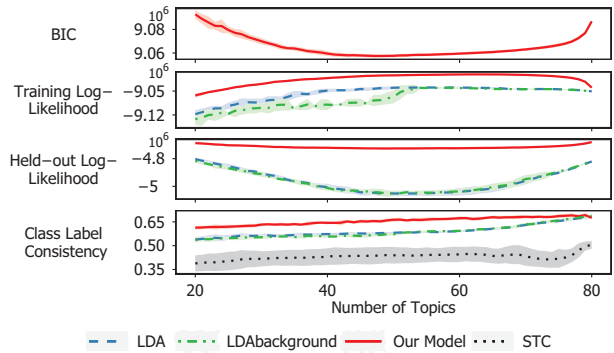}
\vspace{-0.05in}
\caption{Average performance versus number of topics on LabelMe image domain.}
\vspace{-.2in}
\label{img}
\end{figure}

We initialized the models with $80$ topics and reduced the number of topics by one at each step. Fig. \ref{img} shows performance of our model compared with LDA and LDAbackground. The minimum of BIC is on average at $M^*=46$ (std. dev. = $3.4$). Held-out set log-likelihood is higher in our model compared to LDA and LDAbackground, across all model orders. 

We also performed single-label classification, similar to the procedure described for the single-labeled text corpora. Fig. \ref{img} shows the class label consistency of our model compared to LDAbackground, LDA, and STC. We can see that our model achieves better label consistency than LDA, LDAbackground, and STC.

Tables \ref{mbar_tab} and \ref{nbar_tab} compare the sparsity measure in our model with LDA, LDAbackground, and STC for this dataset at $M^*=46$. At this model order, the average number of occurring topics for LDA is $37.15$, which seems implausible, but our model and STC provide more reasonable sparsity in topic proportions. However, unlike STC, our model exhibits parsimony in word probabilities, with on average only $95.73$ words modeled in a topic-specific fashion. 

\textit{Discussion:} On all four data sets, our models achieve higher held-out likelihoods than LDA, at {\it nearly} all model orders, except for high orders on 20-Newsgroups.  Even stronger, our model evaluated at the single (BIC minimum) order achieves higher held-out likelihood than the LDA models evaluated at nearly all orders.  With respect to class label consistency, our BIC-minimum model gives accuracy better than LDA evaluated at {\it nearly} all orders.  However, it is seen on all the data sets that LDA's class consistency improves with increasing order, even exceeding our model's on 20-Newsgroups for $M > 70$. In general, our model has much greater topic sparsity than LDA,
and with only modest increases in sparsity achieved for LDA by using a background model.
Moreover, our model's average number of topics per document has good agreement with the average number of class labels per document.  In this sense, our learned topics better correspond to individual classes than LDA's.  From the curves in Figs. \ref{c}, \ref{b}, \ref{cc}, and \ref{img}, it may be possible for LDA to achieve greater class label consistency than our model on some of the data sets by using a {\it very} large number of topics.  However, one must recognize that topic models perform unsupervised {\it clustering}, aiming to capture the most salient content and to provide human-interpretable data groupings.  Choosing a huge number of topics may improve LDA's class label consistency, but such solutions would defy human interpretation. Our BIC-minimizing solutions are more interpretable due to their topic sparsity, their {\it word} sparsity and, as noted, due to topics better corresponding to ground-truth classes.  Moreover, this parsimony and interpretability are achieved without giving up performance (held-out set likelihood and class consistency). 

\textit{Execution Time:} We report the execution times for the complete model learning process, from $M=M_{\rm max}$ down to $M=M_{\rm min}$, for our model, LDA, and STC on the four datasets in Table \ref{runtime}. All models were run on machines with Quad-Core 3.0 GHz processors. For the datasets with small $\overline{M}$ or modest document length, execution time of our model is smaller or comparable to LDA. But on the Ohsumed corpus with $\overline{M}=1.85$ and larger typical document lengths, running time in our model is three times that of LDA. Also, STC has the smallest overall running time across these datasets. 
\begin{table}[ht]
\caption{Average execution times in minutes (std. dev.)}
\centering
\begin{tabular}{ |c|c|c|c| }
\hline 
Dataset & Our Model & LDA & STC \\ \hline
Ohsumed &	2529 (97)	&	818 (251)	&	243 (11)\\ \hline
20-Newsgroup &	757 (46)	&	1510 (94)	&	527 (10)\\ \hline
Reuters &	395 (31)	&	746 (52)	&	140 (9)\\ \hline
LabelMe &	209 (7)	&	124 (6)	&	280 (30)\\ \hline
\end{tabular}
\label{runtime}
\end{table}

The E-steps in both variational inference for LDA and EM for our model can be readily parallelized and performed separately for each document. This will improve scalability to larger data sets. Nevertheless, we ran the standard implementation of our model and LDA, learning a single model with $M=100$ topics on the NSF Research Awards corpus\footnote{http://kdd.ics.uci.edu/databases/nsfabs/nsfawards.data.html} which consists of 129000 abstracts of NSF awards from 1990-2003. Training time for our model and LDA were respectively 30 and 10 hours on this (much larger) data set.
\section{Conclusion}
\label{conc}
We have proposed a parsimonious model for estimating topics and their salient words, given a database of unstructured documents. Unlike LDA, our model gives sparse representation both in topic presence in documents and in word probabilities under different topics. We have derived a BIC objective function specific to our model, 
with complexity penalization consistent with the effective sample size for each parameter type.
We minimize this objective to jointly determine 
our model, including the total number of topics. Experiments show that our model outperforms LDA and a sparsity-based topic model \cite{Zhu2011} with
respect to several clustering performance measures, including test set log-likelihood and agreement with ground-truth class labels. 


\end{document}